\documentclass{ifacconf}

\usepackage{natbib}        

\usepackage{amsmath,amsfonts, mathtools, amssymb, multirow}
\usepackage{algorithmic}
\usepackage{algorithm}
\usepackage{array}
\usepackage{textcomp}
\usepackage{stfloats}
\usepackage{url}
\usepackage{verbatim}
\usepackage{graphicx}
\usepackage{subcaption, multirow}

\makeatletter
\let\old@ssect\@ssect 
\makeatother

\usepackage{hyperref}

\makeatletter
\def\@ssect#1#2#3#4#5#6{
  \NR@gettitle{#6}
  \old@ssect{#1}{#2}{#3}{#4}{#5}{#6}
}
\makeatother

\hypersetup{
	colorlinks=true,
	linkcolor=blue,
	citecolor=blue,
	urlcolor=blue,
}

\makeatletter
\def\blfootnote#1{\add@tok\t@glob@notes{\begingroup\def\@thefnmark{}\footnotetext{#1}\endgroup}}
\makeatother

\begin{document}
\begin{frontmatter}

\title{Autonomous Vehicle Collision Avoidance With Racing Parameterized Deep Reinforcement Learning} 

\author[First]{Shathushan Sivashangaran} 
\author[Second]{Vihaan Dutta}
\author[Third]{Apoorva Khairnar} 
\author[First]{Sepideh Gohari}
\author[First]{Azim Eskandarian}

\address[First]{Virginia Commonwealth University, 
   Richmond, VA 23284, USA (e-mail: sivashangars@vcu.edu; goharis@vcu.edu; eskandariana@vcu.edu)}
\address[Second]{University of Michigan, Ann Arbor, MI 48109, USA (email: vihdutta@umich.edu)}
\address[Third]{Virginia Tech, Blacksburg, VA 24061, USA (e-mail: apoorvak@vt.edu)}

\begin{abstract}                % Abstract of 50--100 words

Road traffic accidents are a leading cause of fatalities worldwide. In the US, human error causes 94\% of crashes, resulting in excess of 7,000 pedestrian fatalities and \$500 billion in costs annually. Autonomous Vehicles (AVs) with emergency collision avoidance systems that operate at the limits of vehicle dynamics at a high frequency, a dual constraint of nonlinear kinodynamic accuracy and computational efficiency, further enhance safety benefits during adverse weather and cybersecurity breaches, and to evade dangerous human driving when AVs and human drivers share roads. This paper parameterizes a Deep Reinforcement Learning (DRL) collision avoidance policy Out-Of-Distribution (OOD) utilizing race car overtaking, without explicit geometric mimicry reference trajectory guidance, in simulation, with a physics-informed, simulator exploit-aware reward to encode nonlinear vehicle kinodynamics. Two policies are evaluated, a default uni-direction and a reversed heading variant that navigates in the opposite direction to other cars, which both consistently outperform a Model Predictive Control and Artificial Potential Function (MPC-APF) baseline, with zero-shot transfer to proportionally scaled hardware, across three intersection collision scenarios, at 31× fewer Floating Point Operations (FLOPS) and 64× lower inference latency. The reversed heading policy outperforms the default racing overtaking policy in head-to-head collisions by 30\% and the baseline by 50\%, and matches the former in side collisions, where both DRL policies evade 10\% greater than numerical optimal control. 

\end{abstract}


\end{frontmatter}
%===============================================================================

\section{INTRODUCTION} \label{se:introduction}

The safety of Autonomous Vehicles (AVs) is predicated on the ability to maintain navigation integrity within high-stakes edge-cases, where standard motion planning fails. This necessity is underscored by the human and economic toll of modern traffic collisions, where human error remains the primary factor in 94\% of crashes. In recent years, annual pedestrian deaths in the United States reached 7,508, and motor vehicle injury costs were estimated at \$513.8 billion (\cite{farrah2026hittheroad}). While AVs offer a solution to substantially improve road safety, the near future will comprise a mix of AVs and human drivers, thus AVs that possess emergency collision avoidance systems capable of offsetting unpredictable dangerous driving better prevent fatalities.

AVs dependence on sensors and computers provides advantages such as untiring focus and accuracy, while simultaneously necessitating counter measures. Adverse weather conditions (\cite{zhang2023perception}) such as heavy snow, torrential rain, or dense fog impair visibility and decrease time for decision-making. Beyond environmental factors, the increased dependence on computers in the AV stack and traffic infrastructure elevates susceptibility to systemic failures and cybersecurity breaches (\cite{lou2019survey, rajabli2020security}). For instance, a malicious actor may hack signalized intersection controllers to display green lights in all directions, creating a dangerous crossing conflict. In these adversarial scenarios, reactive braking is often insufficient; the vehicle must execute safe, evasive maneuvers at the limits of tire friction to maximize the probability of collision avoidance. Furthermore, these maneuvers are technically similar to strategic interceptor evasion in military applications, where ballistic missiles and Unmanned Aerial and Ground Vehicles evade dynamic interceptors at high velocities, facilitating potential technology transfer across domains.

Executing evasive maneuvers within the narrow temporal window of an impending crash imposes a dual constraint on the vehicle control architecture; the necessity to operate within the nonlinear regime of vehicle physics and the requirement for computational efficiency (\cite{thompson2024adaptive, guo2016nonlinear}). The predominant methods for emergency avoidance comprise optimization planners such as Model Predictive Control (MPC) (\cite{thompson2024adaptive, liu2017combined}), Artificial Potential Functions (APF) (\cite{feng2021collision}) and Control Barrier Functions (CBF) (\cite{ames2016control}). These methods provide a structured framework for multi-objective optimization and vehicle stability, with nonlinear vehicle models. Solving physically accurate, nonlinear optimization matrices requires significant computation. This latency regularly exceeds the real-time processing thresholds required for evasion in sudden path conflicts. Consequently, a paradox emerges where when the models are mathematically simplified to meet time constraints, the collision avoidance success rate is compromised as the controller fails to optimize the vehicle's behavior at its physical limits.

Machine Learning (ML), specifically Deep Learning (DL) and Deep Reinforcement Learning (DRL), offer a resolution to this paradox by shifting the computational overhead from real-time execution to offline training. However, the training samples required often exceed several million state-action pairs that comprise collisions which can be infeasible without synthetic data in simulation. Furthermore, the transfer of simulation trained policies to the physical domain is inherently arduous, especially with instantaneous onboard sensor observations instead of maps and reference positions, as is required for unplanned collision avoidance. This paper presents an AV collision avoidance DRL policy parameterized Out-Of-Distribution (OOD) with race car overtaking using a physics-informed reward instead of reference trajectory aided learning. Collision avoidance success rate is increased by reversing the heading of the trained agent to avoid oncoming high-momentum race cars. The policy is trained in an analytically accurate physics engine over 60,000,000 samples in a 3D rigid-body multi-agent environment, with a simulator exploit-aware reward to transfer zero-shot to hardware. Racing serves as an information-dense proxy where updating the Artificial Neural Network's (ANN's) weights to maximize track performance intrinsically parameterizes the nonlinear kinodynamics required for crash avoidance, as both minimizing lap time and executing emergency evasions necessitate split-second maneuvering at the limits of tire friction.

The rest of this paper is organized as follows. Section \ref{sec_2} summarizes the state-of-the-art in AV collision avoidance, Section \ref{sec_3} presents the method which comprises the DRL formulation, simulation training and hardware transfer, the results are analyzed in Section \ref{sec_4}, and Section \ref{sec_5} provides a conclusion.

\section{RELATED WORK} \label{sec_2}

Collision avoidance frameworks frequently utilize MPC, APF, CBF and risk assessment models to define safe navigable zones. These comprise charged particle representations to repel obstacles, enabling real-time routing through continuous reactive forces (\cite{reichardt1994collision}), predictive occupancy map methods that proactively predict trajectories of surrounding vehicles to select the safest local path within acceleration limits (\cite{lee2019collision}) and probabilistic frameworks that use conditional random fields to estimate collision threats via metrics such as times to collision, stop and evade (\cite{li2021risk}). MPC is widely adopted to account for nonlinear kinematics and actuator limits in emergency scenarios, such as temporarily violating stability criteria at the limits of tire friction to prevent imminent crashes (\cite{funke2016collision}), dynamically activating lane-changing and braking with sigmoid-based convex safety barriers (\cite{ammour2022mpc}) and integrating with APF utilizing a line-charge potential field as the cost function for variable road structures such as intersections, with targeted occupant protection by concentrating charge around the driver's position when a crash becomes mathematically inevitable (\cite{shang2023emergency}).

Nonlinear physics at the vehicle's limits to maximize collision avoidance probability necessitates computationally expensive vehicle models that impact the effectiveness of classical methods in real-time deployment on hardware. Conversely, ML such as DRL parameterize nonlinear dynamics in an ANN with low latency by shifting computation from online inference to offline training. These comprise dedicated dynamic obstacle avoidance architectures that commonly utilize Long Short Term Memory (LSTM) to encode obstacle spatial positions in a fixed-length hidden state for informed kinodynamic planning such as for pedestrian avoidance, and integration of LSTMs with DRL for generalization to UAVs and UGVs (\cite{everett2021collision}). Race AV multi-agent DRL policies trained and validated in open-source simulators collectively optimize collision avoidance and lap time of all agents (\cite{yuan2020race}), prioritizing collision avoidance by incentivizing going off-track, instead of higher risk overtaking, that is beneficial for high velocity evasive maneuvers. 

The growing connectivity of AVs through Vehicle-to-Everything (V2X) communication introduces substantial cybersecurity risks. Attacks may target sensors via GPS spoofing or adversarial LiDAR and camera perturbations (\cite{petit2015remote, cao2019adversarial, sun2020towards}), communication channels through denial-of-service, or infrastructure via falsified V2X messages broadcasting incorrect road conditions (\cite{breitling2021security}). These attacks represent OOD events that extend beyond typical training scenarios, posing unique and ongoing challenges for AV safety. Waymo's collision avoidance testing methodology corroborates a single policy is unlikely to anticipate the full breadth of hazardous edge cases encountered in deployment with scenario databases constructed from naturalistic crash data, on-road testing, and structured combinatorial analysis (\cite{kusano2022collision}), underscoring the necessity for emergency collision avoidance.

\begin{figure*}[!h]
    \centering
    \includegraphics[width = 1.7\columnwidth]{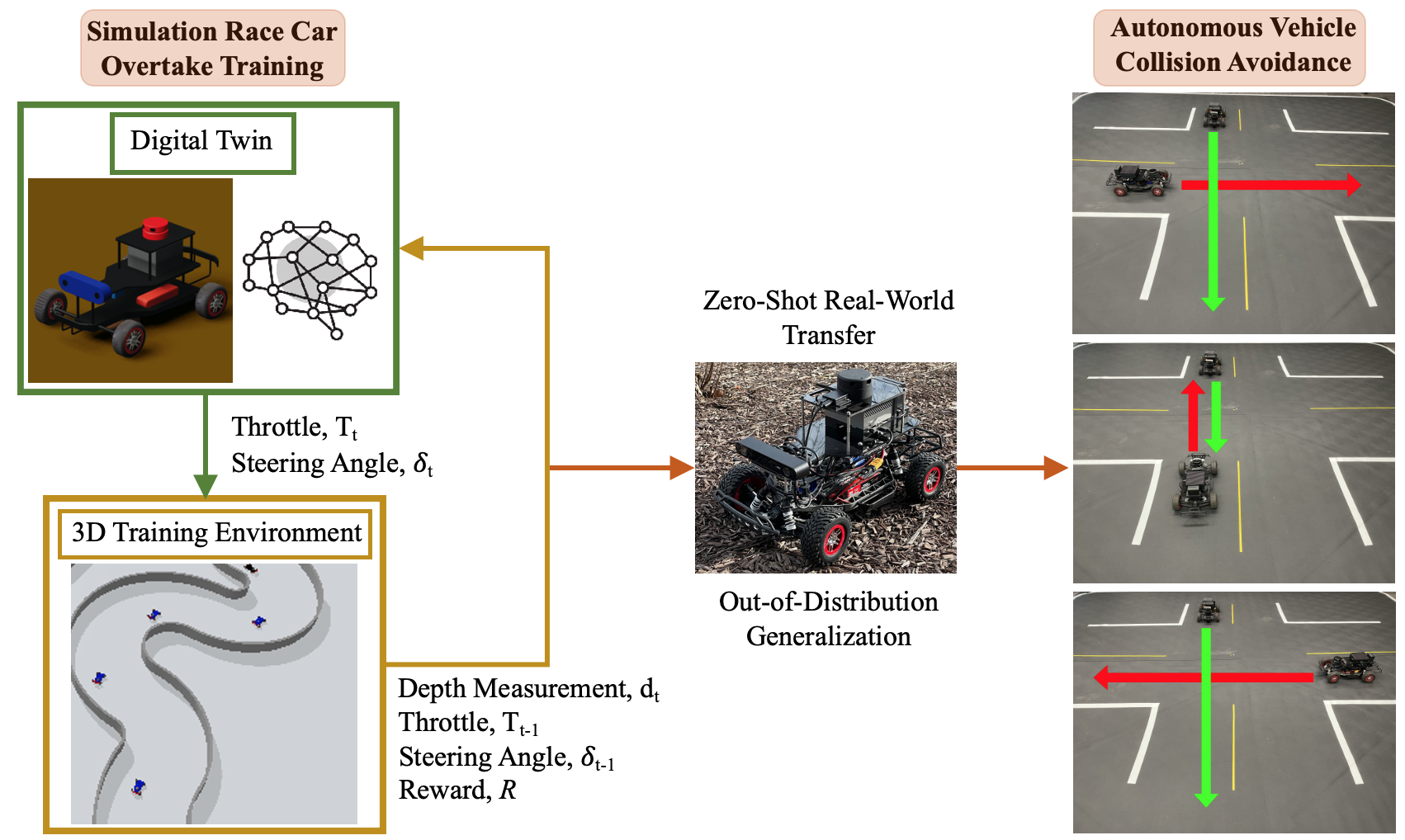}
    \caption{Schematic of the method. The DRL policy is trained in simulation to avoid oncoming race cars while minimizing lap time and transferred zero-shot to $1/10^{th}$ proportionally scaled hardware for OOD AV collision avoidance. A video can be viewed at \href{https://www.youtube.com/watch?v=NQudp6cH-20}{https://www.youtube.com/watch?v=NQudp6cH-20}.} \label{method_schematic}
\end{figure*}

\section{METHODOLOGY} \label{sec_3}

Nonlinear vehicle kinodynamics for dynamic AV collision avoidance is parameterized in a 2 hidden layer Multilayer Perceptron (MLP) with $tanh$ activation and 64 nodes in each hidden layer, with Proximal Policy Optimization (PPO) (\cite{schulman2017proximal}) in a simulator (\cite{sivashangaran2023autovrl}) built on the Bullet physics engine (\cite{coumans2021}), by training to avoid oncoming race cars, while minimizing lap time. The policy is validated for OOD road collision avoidance in an intersection with $1/10^{th}$ proportionally scaled AV hardware (\cite{sivashangaran2023xtenth}), depicted in Figure \ref{method_schematic} which illustrates the method.

\subsection{Observation and Action Spaces}

\textbf{Observations. }Spatial depth rays across a frontal $90^{\circ}$ Field of View (FoV) comprise the observation space $\mathcal{O}$, defined as follows, where $r_{i}$ is the depth measurement at each angle increment within the FoV. 

\begin{equation} \label{}
\mathcal{O} = [r_{i}]^{1\times170}
\end{equation}

170 rays were utilized for the Pareto optimization of training convergence, inference time and performance, determined empirically. The rays were simulated as detailed in (\cite{sivashangaran2023autovrl}). On hardware, both 2D LiDAR and camera RGB-D point clouds were algorithmically shaped to the geometric occupancy of $\mathcal{O}$. Among the evaluated sensors that comprise $360^{\circ}$ 2D LiDARs, the YDLIDAR G2 and RPLIDAR S2, and the Intel RealSense D435i infrared camera, RPLIDAR S2 was used for its seven times greater data density over YDLIDAR G2 that resulted in more accurate object detection, and higher sampling rate over Intel RealSense D435i for high frequency inference.

\textbf{Actions. } Throttle $T$ and steering angle $\delta$ in the range $[\delta_{min}  \delta_{max}]$ comprise controllable actions $a$. These are computed normalized within $[-1\;\;\;1]$ by the DRL algorithm $\textit{a} = (a_{T}, a_{\delta})$, and scaled to vehicle applicable controls as follows.

\begin{equation} \label{}
T = min(max(a_{T}, 0), 1)
\end{equation}

\begin{equation} \label{}
\delta = max(min(a_{\delta}, \delta_{max}), \delta_{min})
\end{equation}

$T$ is nonlinearly related to the vehicle's velocity $v$ with a friction model, as detailed in (\cite{sivashangaran2023autovrl}), which facilitates the self-supervised physics-informed reward to encode spatial density velocity potentials.

\subsection{Simulator-Exploit Aware Reward}

A non-geometric mimicry, physics-informed reward component drives the primary learning objective. This is combined with auxiliary simulator-exploit pruning and the replacement of an explicit collision penalty with an implicit truncation of the value horizon, as detailed in (\cite{sivashangaran2026racing}). The latter and the independence of precomputed reference trajectories facilitate greater OOD generalization by minimizing variance induced conservatism and projecting spatial observations directly to nonlinear vehicle dynamics considerate velocity potentials without geometric positional guidance. 

The reward $R$ is formulated as follows where $5T^{2}$ is the physics-informed component that is proportional to $v^{4}$ which optimizes trajectories that maximize momentum at the limits of the vehicle's nonlinear tire friction circle, and the penalty of $-2.0$ for the product of the previous $a_{\delta,prev}$ and current $a_{\delta,curr}$ steering angles at $-1$, prunes an exploitation of the discrete-time simulator mechanics that permits non-physical minimum to maximum steering reversals, which otherwise results in slaloming on hardware (\cite{evans2023comparing, brunnbauer2022latent}).

\begin{equation} \label{eqn_reward_f}
    R = 
\begin{dcases}
    5T^{2} &\\ 
    -2.0              & \text{if } a_{\delta,prev}a_{\delta,curr} = -1\\
\end{dcases}
\end{equation}

To minimize excessive track boundary and obstacle vehicle avoidance by simultaneously discouraging collisions while maximizing the friction circle, a penalty is not applied upon collision and episode termination. Instead, the value horizon $V(\mathcal{O})$ decreasing to $0$ instead of $\sum_{k=0}^{\infty}\gamma^kT_{}^2$ where $\gamma$ is the discount factor is used as implicit dynamic lost opportunity cost.

\subsection{Training \& Hardware Transfer}

\begin{figure}
\centering
\begin{minipage}{0.252\textwidth}
\includegraphics[width=0.8\textwidth]{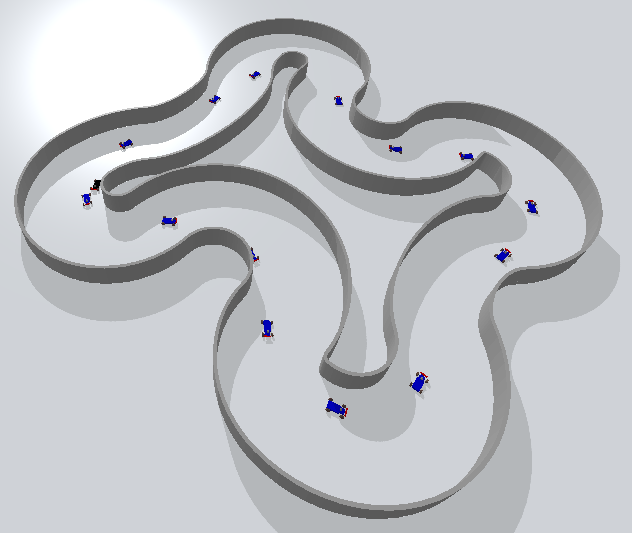}
\subcaption{} \label{train_env_full}
\end{minipage}%
\begin{minipage}{0.252\textwidth}
\includegraphics[width=0.8\textwidth]{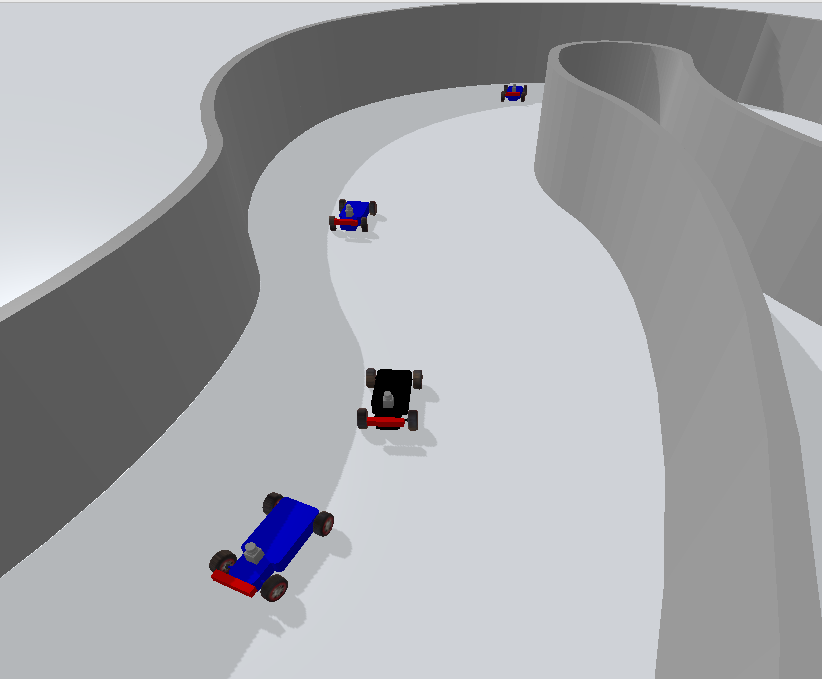}
\subcaption{} \label{train_env_zoom_default}
\end{minipage}
\\
\begin{minipage}{0.252\textwidth}
\includegraphics[width=0.8\textwidth]{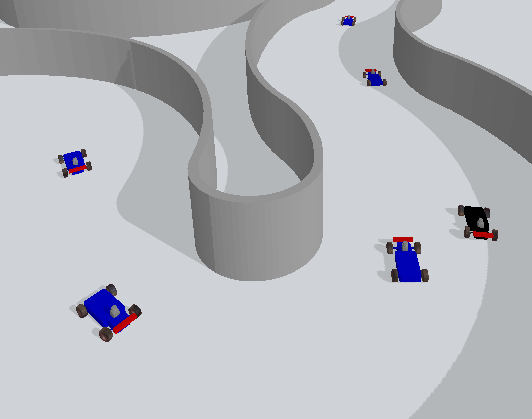}
\subcaption{} \label{train_env_zoom_oncoming}
\end{minipage}%

\caption{Simulation training environment. (a) 15 obstacle cars lap with a pretrained DRL racing policy. (b) The overtaking car in black and obstacle cars in blue in the default uni-direction heading. (c) The overtaking car's heading is reversed to avoid oncoming high-speed traffic to better parameterize OOD road AV collision avoidance.} \label{train_env}
\end{figure}

\textbf{Training. }The simulation training environment contains a racetrack, 15 obstacle cars and the agent being trained, imported to Bullet in Unified Robot Description Format (URDF), illustrated in Figure \ref{train_env}. The obstacle cars lap the track with a time-optimal, high momentum racing DRL policy (\cite{sivashangaran2026racing}). The overtaking car's maximum velocity is increased, and it is set in the same direction as the obstacle cars for default race car overtaking, depicted in Figure \ref{train_env_zoom_default}, and flipped opposite to traffic direction, depicted in Figure \ref{train_env_zoom_oncoming} to evade oncoming race cars. Both policies were trained over 100,000,000 samples, with model checkpoints saved every 5,000,000 utilizing 10 epochs, a batch size of 64, rollout buffer size of 2048, $\gamma$ of 0.99 and learning rate of 0.0003. The best performers, which comprise the policies after 40,000,000 for the identical traffic direction and 60,000,000 for reversed heading, were transferred to hardware for evaluation of OOD AV collision avoidance.

\textbf{Hardware Transfer. }Physical sensors and actuators in the scaled car were utilized to replace the physics engine's simulated counterparts for zero-shot policy transfer. RPLiDAR S2 observations were preprocessed to filter and replace outlier measurements with interpolations, and uniformly downsampled to spatially fit $\mathcal{O}$. The controls computed by the policy were applied to the servo and propulsion motors using a calibrated Electronic Speed Controller (ESC). Collision avoidance was tested in a scaled intersection with an obstacle car approaching from the right, head-on and left.

\section{RESULTS AND DISCUSSION} \label{sec_4}

This section presents the simulation training results for reversed heading race car overtaking and the OOD collision avoidance of both DRL policies evaluated on $1/10^{th}$ proportionally scaled hardware, benchmarked against a MPC-APF baseline (\cite{shang2023emergency}).

\begin{figure*}
    \centering
    \begin{minipage}{0.3\textwidth}
    \includegraphics[width=\textwidth]{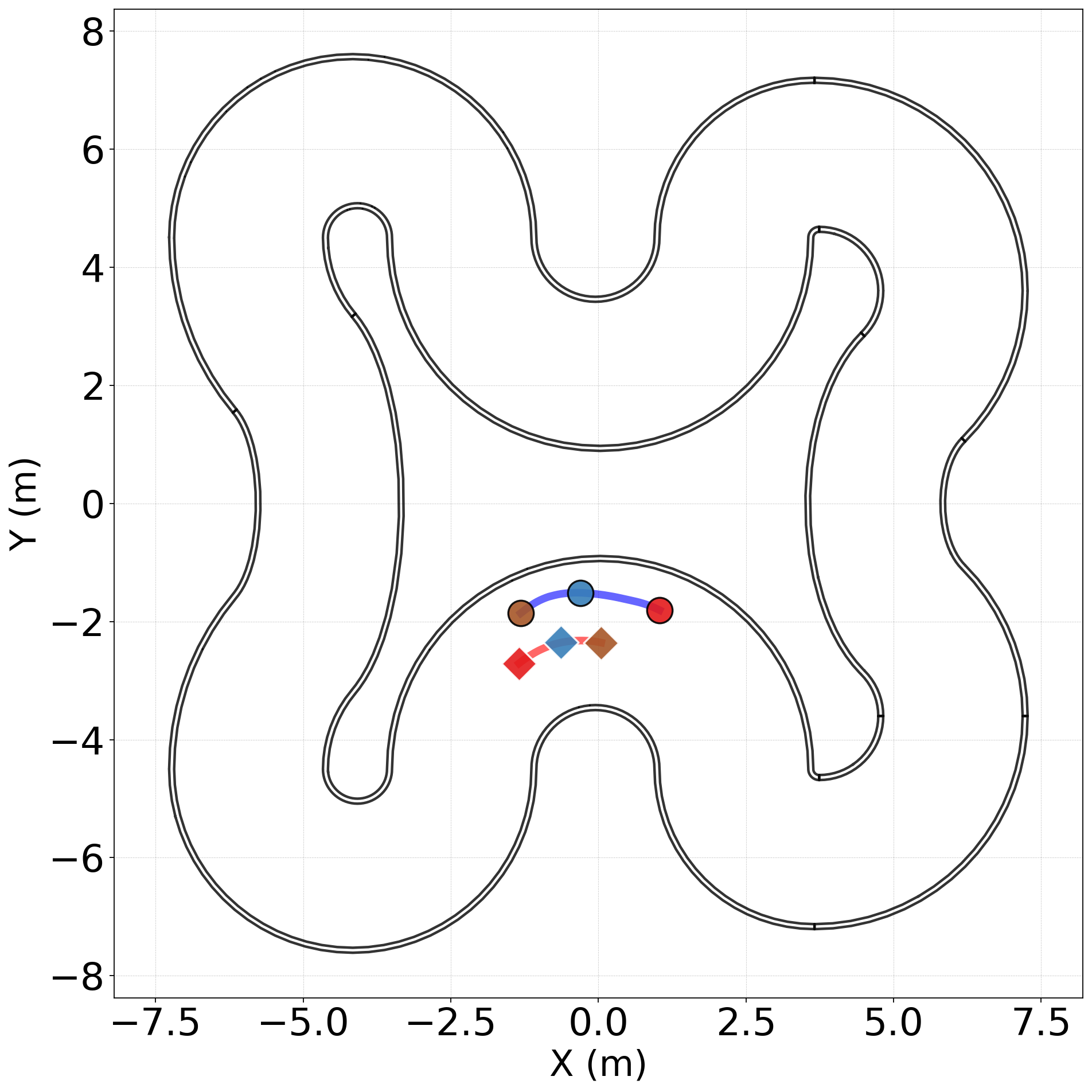}
    \end{minipage}%
    \begin{minipage}{0.3\textwidth}
    \includegraphics[width=\textwidth]{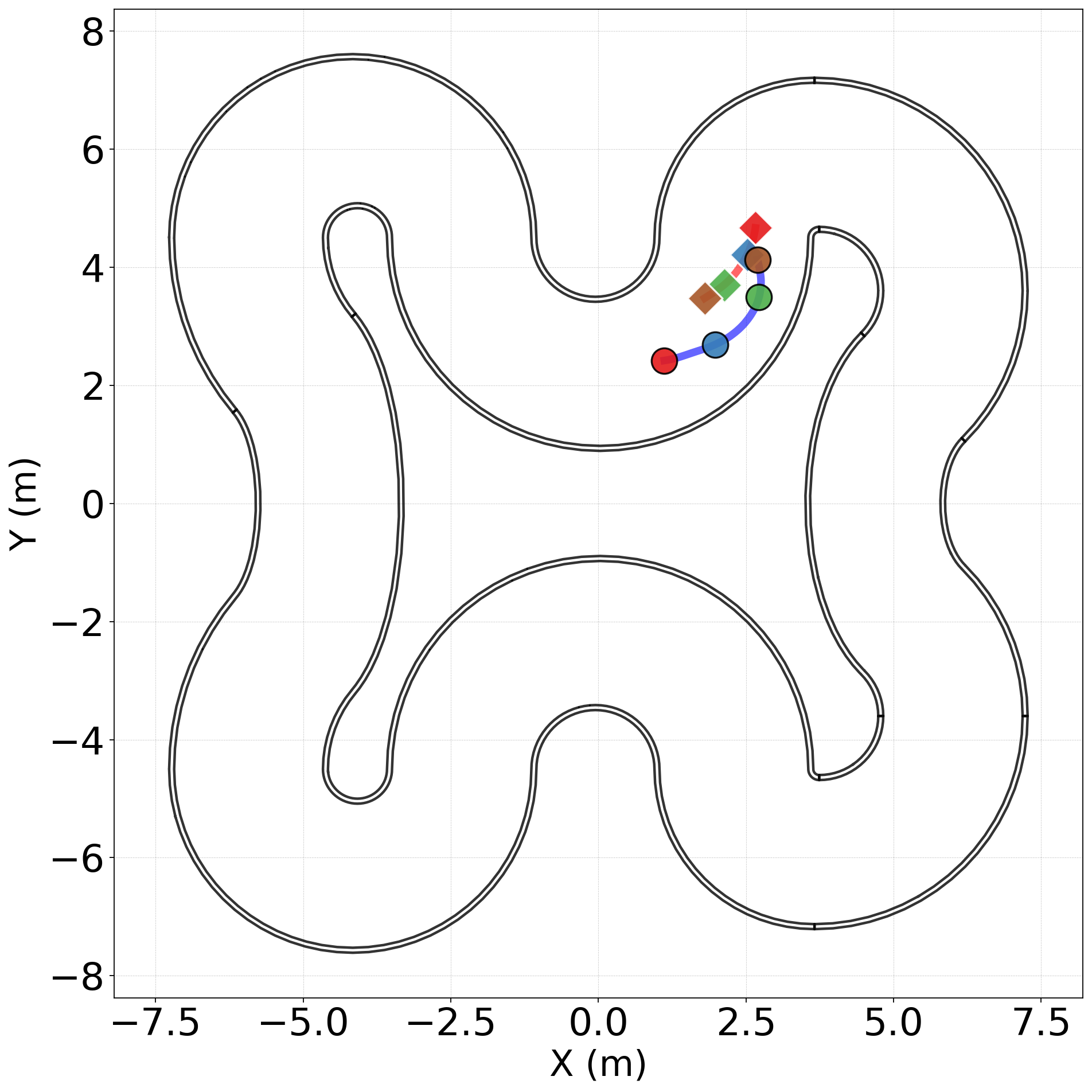}
    \end{minipage}
    \begin{minipage}{0.3\textwidth}
    \includegraphics[width=\textwidth]{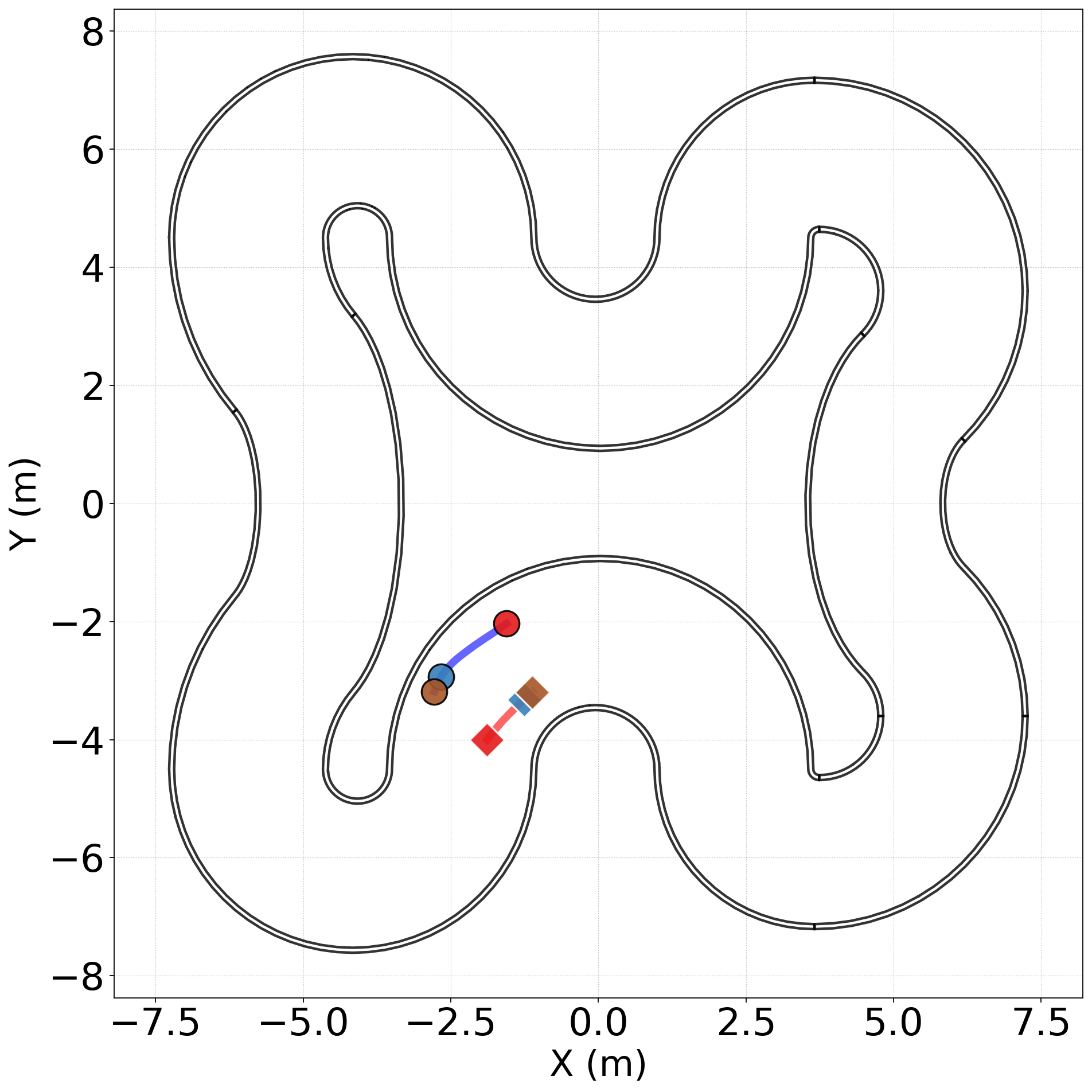}
    \end{minipage}
    \\
    \begin{minipage}{0.4\textwidth}
    \includegraphics[width=\textwidth]{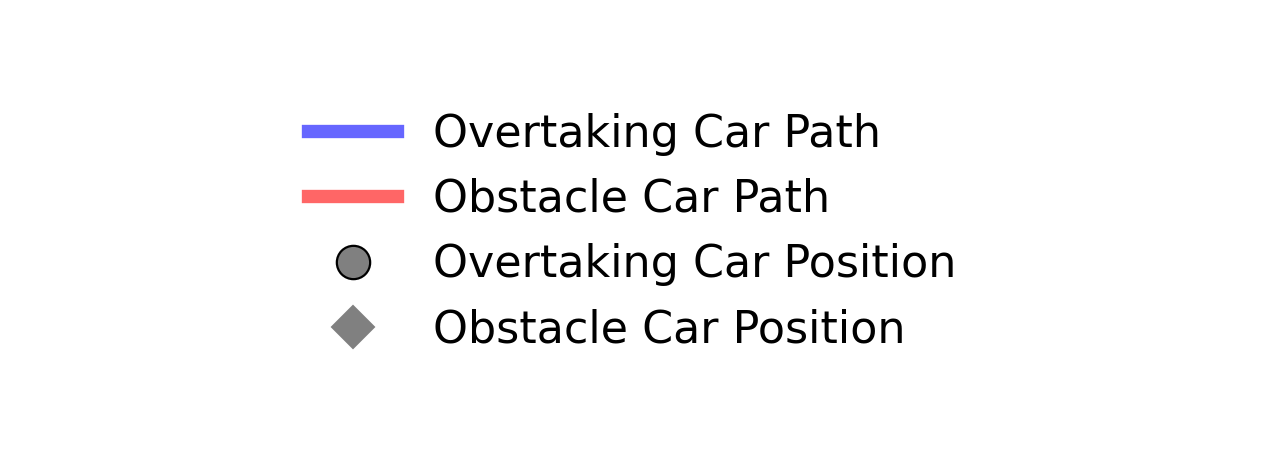}
    \end{minipage}%
    \caption{Trajectories of the overtaking and obstacle cars in the reversed heading simulation training environment. The positions in each trajectory are causally color coded.} \label{plot_ot_traj_oncoming}
\end{figure*}

\textbf{Reversed Heading Race Car Overtaking. }The agent dynamically traverses around high momentum obstacle cars by executing lateral deviations consistent with operation at the tire friction limits. Figure \ref{plot_ot_traj_oncoming} illustrates three example trajectories in the simulation training environment, corresponding to distinct spatial configurations that corroborate the policy generalizes across varying obstacle positions rather than converging to a fixed evasion template.

\begin{figure*}
\centering
\begin{minipage}{0.3\textwidth}
\includegraphics[width=0.9\textwidth]{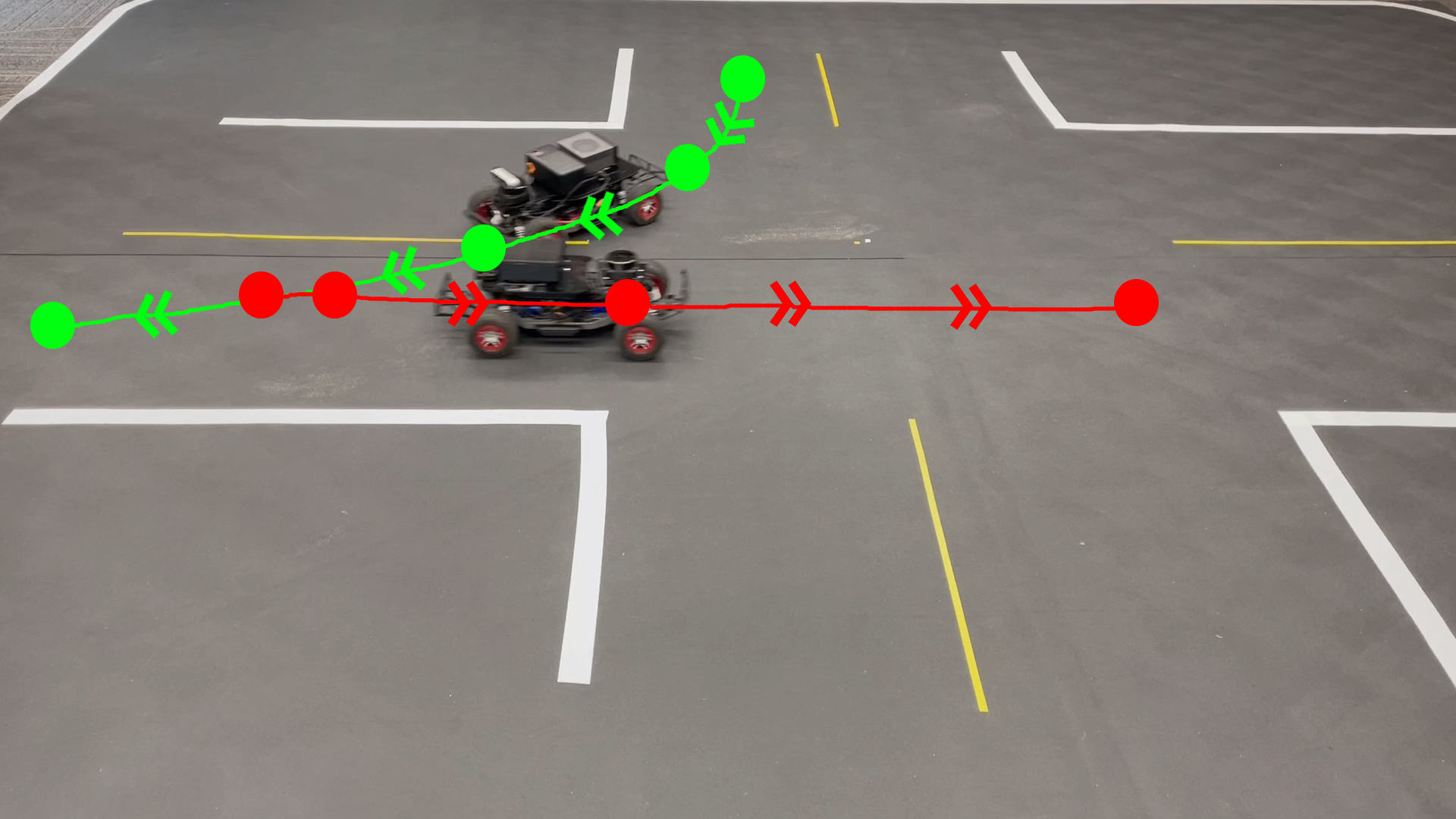}
\subcaption{} \label{traj_1}
\end{minipage}
\begin{minipage}{0.3\textwidth}
\includegraphics[width=0.9\textwidth]{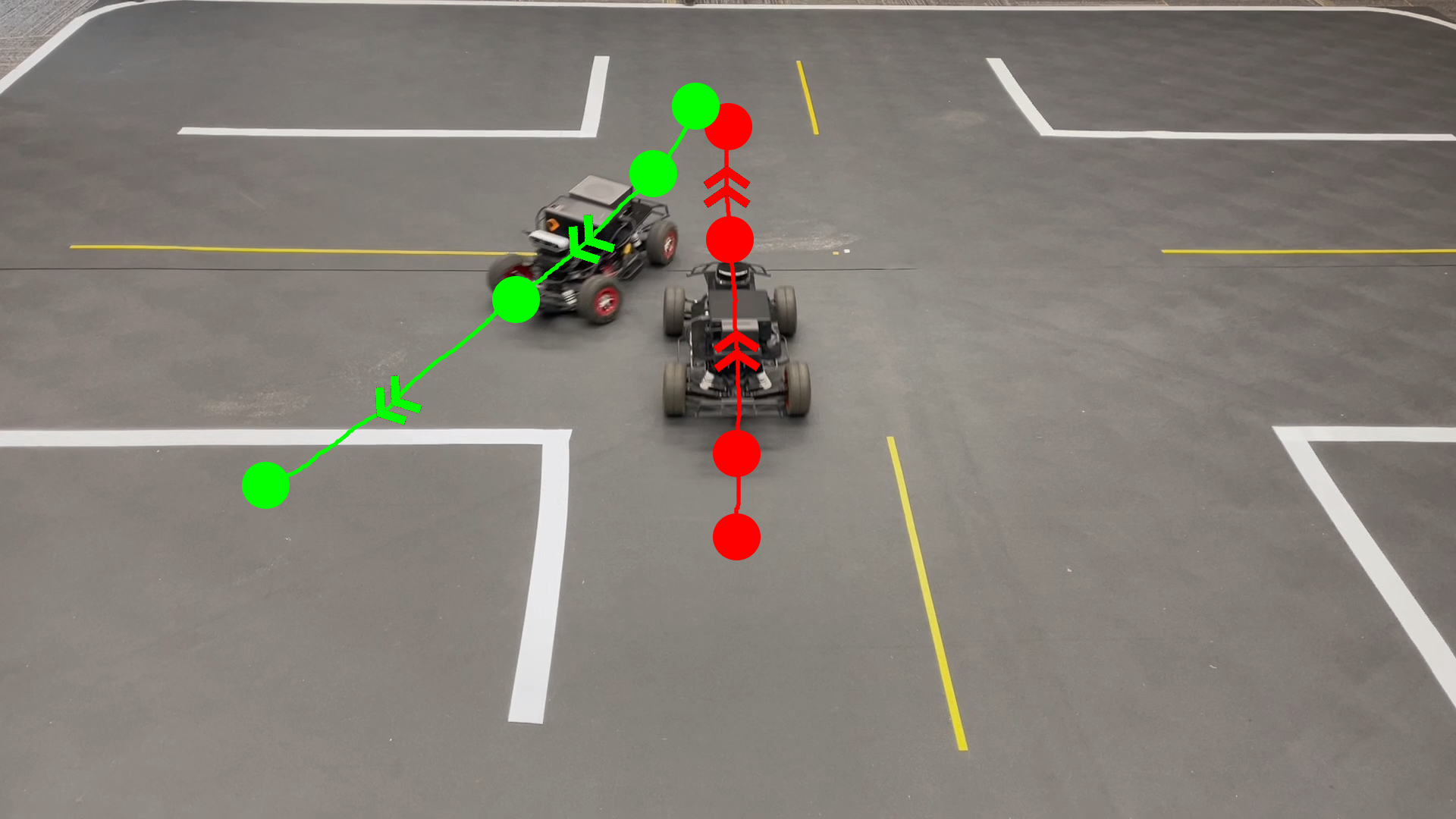}
\subcaption{} \label{traj_2}
\end{minipage}
\begin{minipage}{0.3\textwidth}
\includegraphics[width=0.9\textwidth]{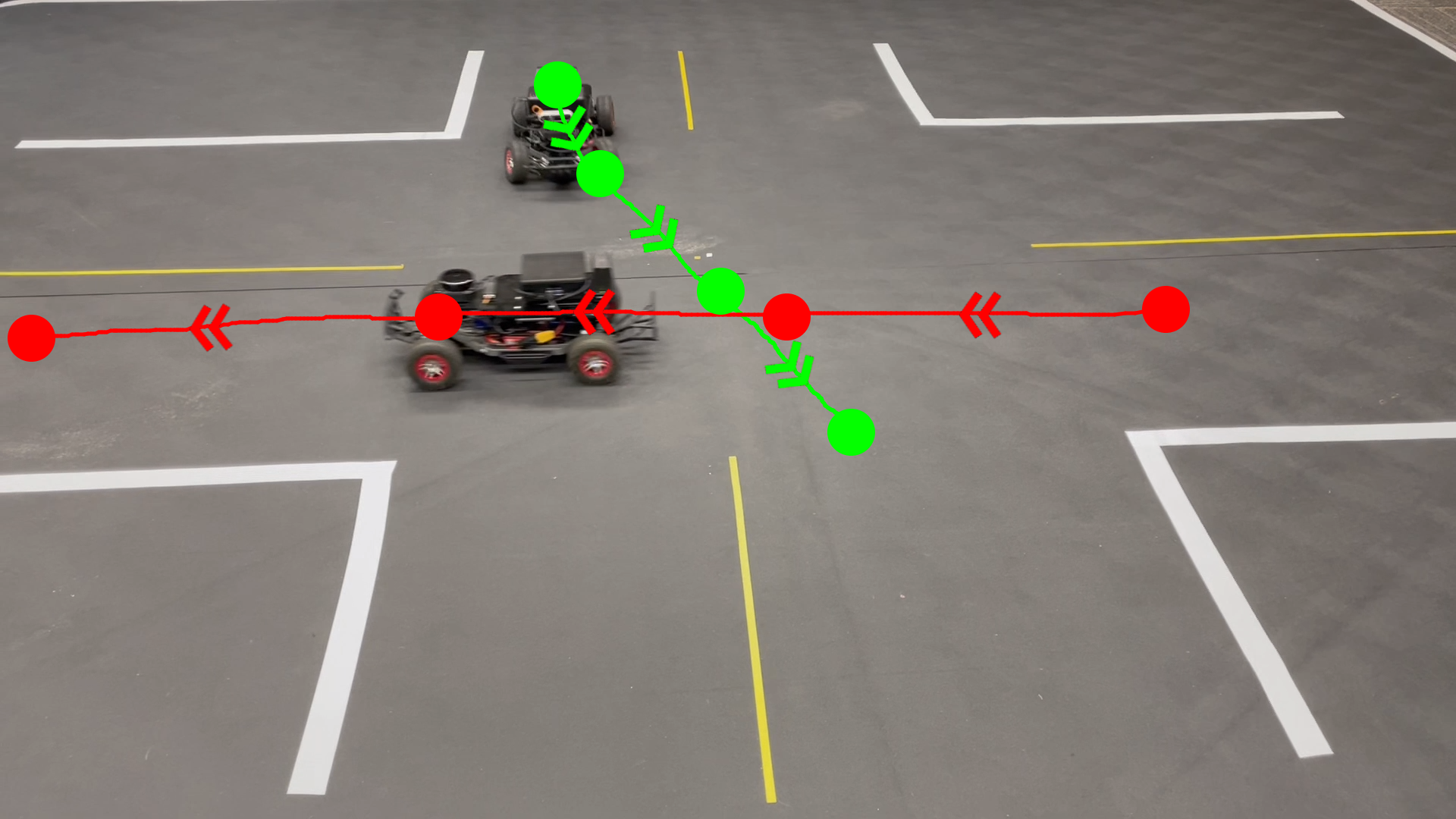}
\subcaption{} \label{traj_3}
\end{minipage}

\caption{Trajectories of the ego car, in green, and obstacle car, in red, on scaled hardware. The obstacle car moves (a) Right-to-Left. (b) Head-to-Head. (c) Left-to-Right.} \label{traj}
\end{figure*}

\textbf{AV Collision Avoidance. }Trajectories of the ego and obstacle vehicles with the reversed heading DRL policy across three intersection collision scenarios, with the obstacle vehicle approaching Right-to-Left (R to L), Head-to-Head (H to H) and Left-to-Right (L to R) are shown in Figure \ref{traj}. Table \ref{coll_avoidance} summarizes the success rates of the two DRL policies and MPC-APF baseline in each scenario at 1 m/s over 10 trials. A linear bicycle model and a horizon of 10 are utilized for the numerical optimization to facilitate a real-time control frequency.

The DRL policies outperform the MPC-APF baseline in all directions, improving by 10\% in side collisions and up to 50\% in H to H, the scenario with the greatest relative velocity. The reversed heading policy performs similarly to the default overtaking policy in side collisions, and doubles avoidance in H to H.

\textbf{Computation Cost. }The Floating Point Operations (FLOPS) and wall-clock inference time on a NVIDIA Jetson Orin AGX for MPC-APF and the DRL ANN are summarized in Table \ref{inference_compute}. The DRL policy requires 30,466 FLOPS and computes control signals in 0.206 $ms$, compared to 960,000 FLOPS and 13.2 $ms$ for MPC-APF which required an average of 12 iterations. This represents a reduction of 31× in FLOPS and 64× in inference time, enabling the DRL policy to operate at significantly higher control frequencies. This computational efficiency is particularly consequential in the H to H scenario, where the narrow temporal window of an oncoming conflict requires sub-millisecond decision latency. 

\begin{table}[ht]
    \renewcommand{\arraystretch}{1.2}
    \centering
    \caption{Collision Avoidance Success Rate (\%)}
    \label{coll_avoidance}
        \resizebox{\columnwidth}{!}{
        \begin{tabular}[t]{|c|c|c|c|c|c|}
            \hline
            \textbf{Method} & \textbf{R to L $\uparrow$} & \textbf{H to H $\uparrow$} & \textbf{L to R $\uparrow$}\\
            \hline
            MPC-APF & 70 & 10 & 60\\
            \hline
            DRL Default  & \textbf{80} & 30 & \textbf{70}\\
            \hline
            \textbf{DRL Reversed Heading} & \textbf{80} & \textbf{60} & \textbf{70} \\
            \hline
        \end{tabular}
        }
\end{table}

\begin{table}[ht]
    \renewcommand{\arraystretch}{1.2}
    \centering
    \caption{Computation Cost}
    \label{inference_compute}
        \begin{tabular}[t]{|c|c|c|}
            \hline
            \textbf{Method} & \textbf{FLOPS $\downarrow$} & \textbf{Time (ms) $\downarrow$}\\
            \hline
            MPC-APF & 960,000 & 13.2 \\
            \hline
            \textbf{DRL ANN}  & \textbf{30,466} & \textbf{0.206} \\
            \hline
        \end{tabular}
\end{table}

\section{CONCLUSIONS} \label{sec_5}

This paper presented a DRL collision avoidance policy for AVs, trained OOD with race car overtaking in simulation, transferred zero-shot to hardware. By eliminating reference trajectory guidance, the physics-informed, simulator exploit-aware reward encoded the relations between spatial observations and velocity potentials in a multi-agent environment, by parameterizing nonlinear vehicle kinodynamics, facilitating generalization to unseen road collision scenarios. Two training configurations were evaluated, a default uni-direction racing overtaking policy, that converged at 40,000,000 samples, and a reversed heading variant that generated dynamic, high-momentum oncoming vehicle training samples representative of road AV conflicts, that took 60,000,000.

Both DRL policies outperformed a MPC-APF baseline across three intersection collision directions on proportionally scaled hardware. In side collision scenarios where the obstacle vehicle approached Right-to-Left and Left-to-Right, both DRL policies avoided 80\% and 70\%, exceeding MPC-APF by 10\% in each. In the Head-to-Head scenario, which had the greatest relative collision velocity necessitating the fastest decision making, the margins were more substantial, where MPC-APF avoided 10\% of collisions, due to the computational latency of iterative optimization, whereas the DRL default overtaking policy improved to 30\%, and the reversed heading policy attained 60\%, 50\% and 30\% greater than MPC-APF and the DRL policy trained in the uni-direction environment. The DRL ANN necessitates 30,466 FLOPS and 0.206 $ms$ on the NVIDIA Jetson Orin AGX per control signal, whereas MPC-APF requires 960,000 FLOPS and 13.2 $ms$, a 31× reduction in compute and 64× reduction in latency that is propitious for real-time evasive control.

Training environment dynamics and transition probabilities are utilized as information-dense proxies for parameterizing AV collision avoidance policies, with the reversed heading variant demonstrating that tactical formulation of traffic dynamics directly transfers to improved OOD performance.

\setcounter{secnumdepth}{0}
\section{Acknowledgments}

This work was funded in part by The Commonwealth Cyber Initiative (CCI).

\bibliography{ifacconf}

\begin{thebibliography}{29}
\providecommand{\natexlab}[1]{#1}
\providecommand{\url}[1]{\texttt{#1}}
\providecommand{\urlprefix}{URL }
\expandafter\ifx\csname urlstyle\endcsname\relax
  \providecommand{\doi}[1]{doi:\discretionary{}{}{}#1}\else
  \providecommand{\doi}{doi:\discretionary{}{}{}\begingroup \urlstyle{rm}\Url}\fi

\bibitem[{Ames et~al.(2016)Ames, Xu, Grizzle, and Tabuada}]{ames2016control}
Ames, A.D., Xu, X., Grizzle, J.W., and Tabuada, P. (2016).
\newblock Control barrier function based quadratic programs for safety critical systems.
\newblock \emph{IEEE Transactions on Automatic Control}, 62(8), 3861--3876.

\bibitem[{Ammour et~al.(2022)Ammour, Orjuela, and Basset}]{ammour2022mpc}
Ammour, M., Orjuela, R., and Basset, M. (2022).
\newblock A mpc combined decision making and trajectory planning for autonomous vehicle collision avoidance.
\newblock \emph{IEEE Transactions on Intelligent Transportation Systems}, 23(12), 24805--24817.

\bibitem[{Breitling et~al.(2021)Breitling, Kupfer, Gabriel, and Eckert}]{breitling2021security}
Breitling, A., Kupfer, T., Gabriel, F., and Eckert, C. (2021).
\newblock Security and privacy issues for connected vehicles.
\newblock \emph{2021 IEEE 18th Annual Consumer Communications \& Networking Conference (CCNC)}, 1--6.

\bibitem[{Brunnbauer et~al.(2022)Brunnbauer, Berducci, Brandst{\'a}tter, Lechner, Hasani, Rus, and Grosu}]{brunnbauer2022latent}
Brunnbauer, A., Berducci, L., Brandst{\'a}tter, A., Lechner, M., Hasani, R., Rus, D., and Grosu, R. (2022).
\newblock Latent imagination facilitates zero-shot transfer in autonomous racing.
\newblock In \emph{2022 International Conference on Robotics and Automation (ICRA)}, 7513--7520. IEEE.

\bibitem[{Cao et~al.(2019)Cao, Xiao, Cyr, Zhou, Park, Rampazzi, Chen, Fu, and Mao}]{cao2019adversarial}
Cao, Y., Xiao, C., Cyr, B., Zhou, Y., Park, W., Rampazzi, S., Chen, Q.A., Fu, K., and Mao, Z.M. (2019).
\newblock Adversarial sensor attack on lidar-based perception in autonomous driving.
\newblock \emph{2019 ACM SIGSAC Conference on Computer and Communications Security}, 2267--2281.

\bibitem[{Coumans and Bai(2016--2021)}]{coumans2021}
Coumans, E. and Bai, Y. (2016--2021).
\newblock Pybullet, a python module for physics simulation for games, robotics and machine learning.
\newblock \url{http://pybullet.org}.

\bibitem[{Evans et~al.(2023)Evans, Jordaan, and Engelbrecht}]{evans2023comparing}
Evans, B.D., Jordaan, H.W., and Engelbrecht, H.A. (2023).
\newblock Comparing deep reinforcement learning architectures for autonomous racing.
\newblock \emph{Machine Learning with Applications}, 14, 100496.

\bibitem[{Everett et~al.(2021)Everett, Chen, and How}]{everett2021collision}
Everett, M., Chen, Y.F., and How, J.P. (2021).
\newblock Collision avoidance in pedestrian-rich environments with deep reinforcement learning.
\newblock \emph{Ieee Access}, 9, 10357--10377.

\bibitem[{Farrah(2026)}]{farrah2026hittheroad}
Farrah, J. (2026).
\newblock Hit the road, mac: The future of self-driving cars.
\newblock Testimony before the U.S. Senate Committee on Commerce, Science, \& Transportation.
\newblock \urlprefix\url{https://www.commerce.senate.gov/services/files/4B417566-2E6B-4460-B38E-1D745F2146C7}.

\bibitem[{Feng et~al.(2021)Feng, Qian, and Wang}]{feng2021collision}
Feng, S., Qian, Y., and Wang, Y. (2021).
\newblock Collision avoidance method of autonomous vehicle based on improved artificial potential field algorithm.
\newblock \emph{Proceedings of the Institution of Mechanical Engineers, Part D: Journal of Automobile Engineering}, 235(14), 3416--3430.

\bibitem[{Funke et~al.(2016)Funke, Brown, Erlien, and Gerdes}]{funke2016collision}
Funke, J., Brown, M., Erlien, S., and Gerdes, J. (2016).
\newblock Collision avoidance and stabilization for autonomous vehicles in emergency scenarios.
\newblock \emph{IEEE Transactions on Control Systems Technology}, 25(4), 1204--1216.

\bibitem[{Guo et~al.(2016)Guo, Hu, and Wang}]{guo2016nonlinear}
Guo, J., Hu, P., and Wang, R. (2016).
\newblock Nonlinear coordinated steering and braking control of vision-based autonomous vehicles in emergency obstacle avoidance.
\newblock \emph{IEEE Transactions on Intelligent Transportation Systems}, 17(11), 3230--3240.

\bibitem[{Kusano et~al.(2022)Kusano, Beatty, Schnelle, Favaro, Crary, and Victor}]{kusano2022collision}
Kusano, K.D., Beatty, K., Schnelle, S., Favaro, F., Crary, C., and Victor, T. (2022).
\newblock Collision avoidance testing of the waymo automated driving system.
\newblock \emph{arXiv preprint arXiv:2212.08148}.

\bibitem[{Lee and Kum(2019)}]{lee2019collision}
Lee, K. and Kum, D. (2019).
\newblock Collision avoidance/mitigation system: Motion planning of autonomous vehicle via predictive occupancy map.
\newblock \emph{IEEE Access}, 7, 52846--52857.

\bibitem[{Li et~al.(2021)Li, Yang, Zhang, Qu, Cao, Cheng, and Li}]{li2021risk}
Li, G., Yang, Y., Zhang, T., Qu, X., Cao, D., Cheng, B., and Li, K. (2021).
\newblock Risk assessment based collision avoidance decision-making for autonomous vehicles in multi-scenarios.
\newblock \emph{Transportation research part C: emerging technologies}, 122, 102820.

\bibitem[{Liu et~al.(2017)Liu, Jayakumar, Stein, and Ersal}]{liu2017combined}
Liu, J., Jayakumar, P., Stein, J.L., and Ersal, T. (2017).
\newblock Combined speed and steering control in high-speed autonomous ground vehicles for obstacle avoidance using model predictive control.
\newblock \emph{IEEE Transactions on Vehicular Technology}, 66(10), 8746--8763.

\bibitem[{Lou et~al.(2019)Lou, Yang, Huo, and Wang}]{lou2019survey}
Lou, X., Yang, Y., Huo, X., and Wang, N. (2019).
\newblock A survey of security threats and countermeasures in autonomous vehicles.
\newblock \emph{2019 IEEE 2nd International Conference on Electronics Technology (ICET)}, 423--428.

\bibitem[{Petit and Shladover(2015)}]{petit2015remote}
Petit, J. and Shladover, S.E. (2015).
\newblock Potential cyberattacks on automated vehicles.
\newblock \emph{IEEE Transactions on Intelligent Transportation Systems}, 16(2), 546--556.

\bibitem[{Rajabli et~al.(2020)Rajabli, Flammini, Noto, and Jafari}]{rajabli2020security}
Rajabli, N., Flammini, F., Noto, M., and Jafari, R. (2020).
\newblock Security challenges of connected and automated vehicles.
\newblock \emph{2020 IEEE International Conference on Smart Computing (SMARTCOMP)}, 73--80.

\bibitem[{Reichardt and Shick(1994)}]{reichardt1994collision}
Reichardt, D. and Shick, J. (1994).
\newblock Collision avoidance in dynamic environments applied to autonomous vehicle guidance on the motorway.
\newblock In \emph{Proceedings of the Intelligent Vehicles' 94 Symposium}, 74--78. IEEE.

\bibitem[{Schulman et~al.(2017)Schulman, Wolski, Dhariwal, Radford, and Klimov}]{schulman2017proximal}
Schulman, J., Wolski, F., Dhariwal, P., Radford, A., and Klimov, O. (2017).
\newblock Proximal policy optimization algorithms.
\newblock \emph{arXiv preprint arXiv:1707.06347}.

\bibitem[{Shang and Eskandarian(2023)}]{shang2023emergency}
Shang, X. and Eskandarian, A. (2023).
\newblock Emergency collision avoidance and mitigation using model predictive control and artificial potential function.
\newblock \emph{IEEE Transactions on Intelligent Vehicles}, 8(5), 3458--3472.

\bibitem[{Sivashangaran and {Eskandarian}(2023)}]{sivashangaran2023xtenth}
Sivashangaran, S. and {Eskandarian}, A. (2023).
\newblock Xtenth-car: A proportionally scaled experimental vehicle platform for connected autonomy and all-terrain research.
\newblock In \emph{ASME International Mechanical Engineering Congress and Exposition}, volume 87639, V006T07A068. American Society of Mechanical Engineers.

\bibitem[{Sivashangaran et~al.(2023)Sivashangaran, Khairnar, and Eskandarian}]{sivashangaran2023autovrl}
Sivashangaran, S., Khairnar, A., and Eskandarian, A. (2023).
\newblock Autovrl: A high fidelity autonomous ground vehicle simulator for sim-to-real deep reinforcement learning.
\newblock \emph{IFAC-PapersOnLine}, 56(3), 475--480.

\bibitem[{Sivashangaran et~al.(2026)Sivashangaran, Khairnar, Gohari, Dutta, and Eskandarian}]{sivashangaran2026racing}
Sivashangaran, S., Khairnar, A., Gohari, S., Dutta, V., and Eskandarian, A. (2026).
\newblock Physics-informed reinforcement learning of spatial density velocity potentials for map-free racing.
\newblock \emph{arXiv preprint arXiv:2604.09499}.

\bibitem[{Sun et~al.(2020)Sun, Cao, Chen, and Mao}]{sun2020towards}
Sun, J., Cao, Y., Chen, Q.A., and Mao, Z.M. (2020).
\newblock Towards robust lidar-based perception in autonomous driving: General black-box adversarial sensor attack with principled defenses.
\newblock \emph{2020 ACM SIGSAC Conference on Computer and Communications Security}, 877--894.

\bibitem[{Thompson et~al.(2024)Thompson, Dallas, Goh, and Balachandran}]{thompson2024adaptive}
Thompson, M., Dallas, J., Goh, J.Y., and Balachandran, A. (2024).
\newblock Adaptive nonlinear model predictive control: Maximizing tire force and obstacle avoidance in autonomous vehicles.
\newblock \emph{IEEE Transactions on Field Robotics}, 1, 318--331.

\bibitem[{Yuan et~al.(2020)Yuan, Tasik, Adhatarao, Yuan, Liu, and Fu}]{yuan2020race}
Yuan, Y., Tasik, R., Adhatarao, S.S., Yuan, Y., Liu, Z., and Fu, X. (2020).
\newblock Race: Reinforced cooperative autonomous vehicle collision avoidance.
\newblock \emph{IEEE transactions on vehicular technology}, 69(9), 9279--9291.

\bibitem[{Zhang et~al.(2023)Zhang, Carballo, Yang, and Takeda}]{zhang2023perception}
Zhang, Y., Carballo, A., Yang, H., and Takeda, K. (2023).
\newblock Perception and sensing for autonomous vehicles under adverse weather conditions: A survey.
\newblock \emph{ISPRS Journal of Photogrammetry and Remote Sensing}, 196, 146--177.

\end{thebibliography}

\end{document}